\documentclass[letterpaper, 10 pt, conference]{ieeeconf}  

\IEEEoverridecommandlockouts                              

\usepackage{amsmath} 
\usepackage{amssymb}  

\usepackage{amsfonts}
\usepackage{algorithmic}
\usepackage{graphicx}
\usepackage{textcomp}
\usepackage{xcolor}
\usepackage[whole]{bxcjkjatype}

\makeatletter
\let\NAT@parse\undefined
\makeatother
\usepackage[colorlinks,bookmarks=true]{hyperref}
\usepackage{booktabs}
 \usepackage{array,multirow}
\usepackage{caption,subcaption}
\usepackage{verbatim}
\usepackage[flushleft]{threeparttable}
\usepackage{colortbl}
\usepackage{cite}
\usepackage{hyperref}

\title{\LARGE \bf
Visual Imitation Learning of Non-Prehensile Manipulation 
Tasks with Dynamics-Supervised Models}

\author{Abdullah Mustafa$^{1}$, Ryo Hanai$^{1}$, Ixchel G. Ramirez-Alpizar$^{1}$, Floris Erich$^{1}$, \\ Ryoichi Nakajo$^{1}$, Yukiyasu Domae$^{1}$ and Tetsuya Ogata$^{2}$
\thanks{$^{1}$A. Mustafa, R. Hanai, I. Ramirez-Alpizar, F. Erich, R. Nakajo, and Y. Domae are with the National Institute of Advanced Industrial Science and Technology (AIST), Japan
        {\tt\small \{am-mustafa, ryo.hanai, ixchel-ramirezalpizar, floris.erich, ryoichi-nakajo, domae.yukiyasu\}@aist.go.jp}}%
\thanks{$^{2}$T. Ogata is with the Graduate School of Fundamental Science and Engineering, Waseda University, Tokyo 169-8555, Japan and also with the National Institute of Advanced Industrial Science and Technology (AIST), Japan
        {\tt\small ogata@waseda.jp}}%
}

\begin{document}

\maketitle
\thispagestyle{empty}
\pagestyle{empty}

\begin{abstract}
Unlike quasi-static robotic manipulation tasks like pick-and-place, dynamic tasks such as non-prehensile manipulation pose greater challenges, especially for vision-based control. Successful control requires the extraction of features relevant to the target task. In visual imitation learning settings, these features can be learnt by backpropagating the policy loss through the vision backbone. Yet, this approach tends to learn task-specific features with limited generalizability. Alternatively, learning world models can realize more generalizable vision backbones. Utilizing the learnt features, task-specific policies are subsequently trained. Commonly, these models are trained solely to predict the next RGB state from the current state and action taken. But only-RGB prediction might not fully-capture the task-relevant dynamics. In this work, we hypothesize that direct supervision of target dynamic states (Dynamics Mapping) can learn better dynamics-informed world models. Beside the next RGB reconstruction, the world model is also trained to directly predict position, velocity, and acceleration of environment rigid bodies. To verify our hypothesis, we designed a non-prehensile 2D environment tailored to two tasks: ``Balance-Reaching" and ``Bin-Dropping". When trained on the first task, dynamics mapping enhanced the task performance under different training configurations (Decoupled, Joint, End-to-End) and policy architectures (Feedforward, Recurrent). Notably, its most significant impact was for world model pretraining boosting the success rate from 21\% to 85\%. Although frozen dynamics-informed world models could generalize well to a task with in-domain dynamics, but poorly to a one with out-of-domain dynamics.  Code available at: \url{https://github.com/Automation-Research-Team/DynamicsMapping-2D}
\end{abstract}


\section{Introduction}

You are holding a tray with a bottle on top. How can you move it around as efficiently as possible without dropping it? For one, you have to always keep an eye on it as you navigate. Upon observing some erroneous motion (sliding or tipping), you must adjust your trajectory to ensure stability. Even for humans, such non-prehensile (no grasping) manipulation \cite{nonprehensileSurvey} poses a significant challenge. In contrast to quasi-static tasks \cite{quasiStatic} (non-prehensile pushing or prehensile grasping), our motivating example requires more dynamic manipulation, ensuring minimal (compounding) errors and a robust recovery policy. Previously, with expert knowledge, such policies were attained through state-based controllers \cite{keepitupright}\cite{nonprehensileICRA}. Yet, our interest lies in exploring a vision-based data-driven approach. Such an approach can reduce the need for specialized state estimation and tracking hardware and software. Also, it would do without simplified mathematical models, expert knowledge, or approximate cost functions.

Recently, vision-based imitation learning approaches have been successfully applied to long-horizon, fine-manipulation, and in-the-wild tasks \cite{chi2024universal}\cite{chi2023diffusionpolicy}\cite{zhao2023learning}. However, our target dynamic manipulation tasks present significant challenges under randomized conditions. We hypothesize that current approaches prompt the vision backbone to prioritize position/shape-related features. While this may suffice for quasi-static manipulation, it may be inadequate with dynamic scenarios. 

In Reinforcement learning settings \cite{RL}, higher-order dynamical states such velocity and acceleration are essential components of the observation space to learn successful dynamic policies. However, adaptation of such dynamical states into an imitation learning framework has not been studied. To address this gap, we propose introducing ``Dynamics Mapping" as an additional training objective compelling the vision backbone to also consider dynamic features such as velocity and acceleration. 

``Dynamics Mapping" can be integrated with End-to-End learning frameworks \cite{levine_end--end_2016}\cite{suzuki2023deep}, as well as pretraining more general world models \cite{ha2018worldmodels}\cite{embed2control}\cite{dreamerv3} (for its improved generalization). In either case, more robust policies are achieved by resolving the latent features ambiguity arising from relying solely on policy loss or RGB observation reconstruction loss. The dynamics-informed latent can both better imitate the dynamic trajectory as well as generalize to new tasks with similar dynamics.

In this work, we validate our approach in a challenging 2D non-prehensile setting on two main tasks. First, we train our world model on one task where significant improvement was attained. Subsequently, the frozen (will not be further trained) world model is adapted to another task with in-domain (similar) dynamics showing good generalization. Conversely, adaptation to tasks with out-of-domain dynamics showed inadequate generalization.

The first contribution of this work is the proposal of ``Dynamics Mapping", a method designed to encourage dynamics-informed latent learning from visual inputs (Section \ref{dyna_map}). Secondly, we have developed a dynamic non-prehensile setting, which could serve as a benchmark for vision-based imitation learning (Section \ref{task_desc}). Lastly, we verified our approach superiority through comparative analysis with different baselines, demonstrating enhanced task performance and generalization (Section \ref{results}).

\section{Related Work}
Our work addresses the potential of learning dynamics-informed \textbf{world models} as a backbone for \textbf{visual imitation learning}, enhancing performance in dynamic tasks, particularly \textbf{non-prehensile manipulation}. This section provides a concise overview of these key concepts. 

\textbf{Visual Imitation Learning.}
Imitation learning is a powerful data-driven learning paradigm, yielding impressive outcomes as demonstrated in RT-1 \cite{rt12022arxiv}. Contrasting state-based learning, which utilizes low-dimensional input, learning directly from high-dimensional image data presents greater complexity. Nevertheless, images are more easily obtained while entailing rich scene information. In visuomotor control, the predominant model architecture integrates a visual encoder backbone—such as CNN (Convolutional Neural Network) or ViT (Vision Transformer)—with a low-level control policy, including MLP (Multi-Layer Perceptron), LSTM (Long Short-Term Memory), or Transformer models. With the shared objective of maximizing expert action likelihood for a given observation, various learning approaches may be adopted. Both the encoder and the policy can be trained jointly end-to-end by back-propagating the policy loss through both modules \cite{levine_end--end_2016}\cite{rt12022arxiv}. Despite its simplicity, end-to-end approaches may suffer from reduced sample efficiency and limited generalization. Incorporating future observation reconstruction as an auxiliary loss \cite{suzuki2023deep} can stabilize learning and enhance sample efficiency. However, their proposed architecture predicts future states based solely on current state without action-based conditioning.  Consequently, its internal transition model may over-fit the training dataset and lack the required level of generalization. 

\textbf{World Modeling.}
Towards improved generalization, world models can be trained for both latent encoding and as transition models. Typically, world models are trained independently of any task-specific policy. They can be trained to be as general as needed by training on a large dataset such as GAIA-1, which comprises 4,700 hours of autonomous driving \cite{hu2023gaia1}. Upon training, such models can be integrated with various control policies such as optimal control \cite{embed2control}, reinforcement learning \cite{ha2018worldmodels}\cite{dreamerv3}, and offline reinforcement learning \cite{demoss2023ditto}. Embed2Control \cite{embed2control} learns a world model with linearized dynamics for next-state prediction. The model is then integrated with an iLQR (iterative Linear Quadratic Regulator) control policy that optimizes an approximate quadratic reward. Recurrent World Models \cite{ha2018worldmodels} independently learn a vision encoder (based on VAE (Variational Auto-Encoder), then model the transition dynamics with a recurrent model, and finally optimize a shallow policy. Decoupling the vision encoder and transition model might fail to encode relevant features from complex scenes without sufficient temporal context. Dreamer(v3) \cite{dreamerv3} avoids this by jointly training the encoder and an RSSM (recurrent state space model), resulting in state-of-the-art performance. Using an offline dataset, DITTO \cite{demoss2023ditto} uses the world model to run an on-policy RL optimization algorithm (REINFORCE), utilizing the policy loss as a distance-based reward. 

Compared to reinforcement learning, the application of world models in imitation learning settings has not been widely investigated, likely because the expert demonstrations dataset provides a sufficient training signal to optimize the vision-backbone/policy jointly. In such scenarios, training a world model independently might not be fully justified without a pressing need for generalization. Additionally, an independently-trained world model might not perform as well, being task-agnostic. Our work aims to evaluate the use of world models in imitation learning settings and provides the necessary adjustments to address more challenging dynamic tasks.  

\textbf{Non-prehensile Manipulation.}
The dynamic tasks of interest encompass the non-prehensile manipulation, where objects are carefully controlled by considering dynamic interactions. Existing works typically employ open-loop control using state-based optimal controllers \cite{keepitupright}\cite{nonprehensileICRA}. In \cite{keepitupright}, the objective was to navigate the robot end-effector to a target pose while balancing an object on a tray. A model predictive controller (MPC) was utilized to plan a trajectory that minimizes the distance to the goal while satisfying balancing constraints. In \cite{nonprehensileICRA}, sequences of motion primitives were identified for a desired motion, and these motions were executed using an LQR controller based on linearized dynamics. In this work, we aim to implement closed-loop vision-based control. We simulated two tasks with analogous objectives of balance-reaching and object tossing. Our motivating goal is reproducing these tasks in the real-world utilizing vision-based control.

\section{Dynamics Mapping}\label{dyna_map}
This study addresses vison-based imitation learning of manipulation tasks that involve dynamic motions, specifically focusing on simulated 2D non-prehensile manipulation tasks. 

\textbf{Problem Setup.} Given a set of expert demonstrations $D=\{\tau_0,\tau_1, \cdots, \tau_n\}$, where each trajectory $\tau_i=\{(I_t, [P_t, V_t, A_t], x_t, a_t)\}_{t=0}^T$ includes the raw visual observations $I_t$, other state information (i.e.\ proprioceptive) $x_t$, and the expert action $a_t$. Additionally, the per-object dynamics state comprises the rigid object's COM (center of mass) position $P_t$, velocity $V_t$, and acceleration $A_t$. The primary objective is to validate the significance of incorporating those dynamics states into the learning process.

\subsection{Model Architecture}
\begin{figure}[tbp]
\centerline{\includegraphics[width=\columnwidth]{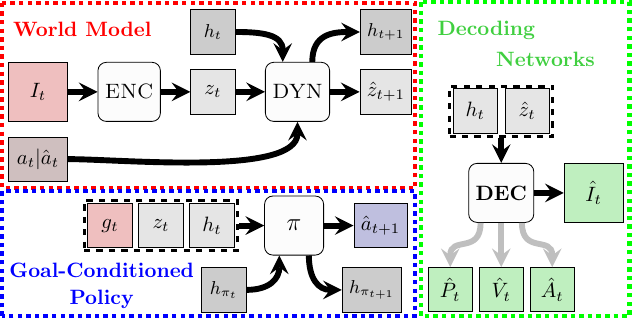}}
\caption{The overall model architecture constitutes a world model, a policy, and a set of decoding networks. The model inputs include RGB state $I_t$, action $a_t$, and goal $g_t$. The internal states include latent $z_t$ and hidden states $h_t$ and $h_{\pi_t}$}. The outputs include predicted action $\hat{a}_t$, dynamics $[\hat{P}_t,\hat{V}_t,\hat{A}_t]$, and reconstructed RGB $\hat{I}_t$.
\label{fig:modelarch}
\end{figure}

The overall model architecture, depicted in Fig. \ref{fig:modelarch}, comprises three main components: a world model, a policy, and a set of decoding networks. The world model's fundamental objective is to encode the input RGB image $I_t$ into a lower-dimensional latent state $z_t$, capturing the necessary environment state for precise action prediction. Towards that, a simple CNN-based encoder ($\textsc{ENC}_{\phi_1}$, parameterized by $\phi_1$) is used for latent encoding ($z_t=\textsc{ENC}_{\phi_1}(I_t)$). However, to tackle sequential tasks with non-Markovian dynamics (reliance on past states), the world model incorporates a transition dynamics model with a recurrent module—comprising two sequential LSTM cells—to preserve the memory essential for dynamics prediction and long-horizon tasks execution. Given the encoded latent $z_t$, current LSTM hidden states ($h_{t} = (h_{1_t}, h_{2_t})$), and the commanded action $a_t$, the transition model ($\textsc{DYN}_{\phi_2}$, parameterized by $\phi_2$) predicts the next-state latent $\hat{z}_{t+1}$ and updates the hidden states ($\hat{z}_{t+1},h_{t+1}=\textsc{DYN}_{\phi_2}(z_t,a_t,h_{t})$). The outputs from the transition module can then be employed for decoding environmental states or predicting future actions.

While the world models capture the general transition dynamics, a task-specific policy is trained based on those world model states towards a target objective. One choice for the goal-conditioned policy is a simple Feedforward MLP architecture (i.e. Dreamer \cite{dreamerv3}). For a goal $g_t$ and a latent state $z_t$, the action is calculated as in Equation \eqref{Feedforward_pi}. 

\begin{equation}
    \hat{a}_{t}=\pi_\theta(z_t,h_t,g_{t})
 \label{Feedforward_pi}
\end{equation}

However, such simple policy cannot handle non-Markovian transitions without dynamics-informed latent features. Analogous to the transition module, a recurrent policy $\pi_\theta$—incorporating two additional LSTM cells—predicts the action $\hat{a}_t$ given the goal position $g_t$, the encoded latent $z_t$, the latest world model hidden states $h_t$, and its own internal states ($h_{\pi_t} = (h_{\pi_{1_t}}, h_{\pi_{2_t}}$) as in Equation \eqref{Recurrent_pi}.

\begin{equation}
\hat{a}_{t},h_{\pi_{t+1}}=\pi_\theta(z_t,g_t,h_{t},h_{\pi_t})
 \label{Recurrent_pi}
\end{equation}

While the world model and the policy could be trained using only the policy loss (refer to the following section), this approach can be sample-inefficient and task-specific. Alternatively, a set of decoding networks are introduced, providing the world model an additional learning signal. Typically, only an unsupervised RGB reconstruction loss is employed; however, we propose introducing supervised dynamics prediction losses for improved performance. The decoders input comprises the encoded latent $\hat{z}_t$ and world model hidden states ($h_{t}$). For RGB decoding, a simple CNN decoder ($\textsc{DEC}_{\zeta_1}$ with weights $\zeta_1$) is utilized ($\hat{I}_t=\textsc{DEC}_{\zeta_1}(\hat{z}_t,h_{1_t},h_{2_t})$). Regarding dynamics decoding, a set of MLP networks (with weights $\zeta_2$,$\zeta_3$,$\zeta_4$) are employed ($[\hat{P},\hat{V},\hat{A}]_t = \textsc{DEC}_{\zeta_{[2,3,4]}}(\hat{z}_t,h_{1_t},h_{2_t})$). The dynamics decoder share the input decoding layer (2-layers MLP) followed by state specific decoding head.

\subsection{Model Training Approaches}
In the literature, two primary approaches for training the discussed model architecture are identified: \textbf{End-to-End} training and \textbf{joint} training. In this work, we have also introduced a \textbf{decoupled} training method. For both joint and decoupled training scenarios, we suggest the incorporation of \textbf{supervised dynamics loss} to learn dynamics-informed models.

\textbf{End-to-End Training.} Given a task-specific dataset $D$, an end-to-end (E2E) training approach is commonly adopted \cite{levine_end--end_2016}. The world model and policy are jointly optimized by minimizing the policy loss as in Equation \eqref{end2endloss_eq}. The loss is computed as the L1-norm between the true and predicted actions. Although straightforward, this approach might be sample-inefficient, necessitating larger datasets for improving performance. Moreover, such task-specific world model might exhibit limited generalization across diverse tasks.

\begin{equation}
 L_{\textrm{E2E}}(\theta,\phi_{1,2}) = E_{D}\big[\|a_t - \hat{a}_t(\theta, \phi_{1,2}) \|\big]
\label{end2endloss_eq}
\end{equation}

\textbf{Joint Training.} 
Besides the policy loss, the RGB reconstruction loss over the next-state observation can serve as an auxiliary optimization objective \cite{suzuki2023deep}. This approach can enhance task-specific performance and capture the underlying environment dynamics. Alongside the discussed policy loss (denoted next as $L_{\pi}$), a cyclic reconstruction loss (Equation \eqref{J_rgbLoss_eq}) is utilized incorporating a weighted sum of the next RGB reconstruction loss ($\|I_{t} - \hat{I}_{t}(\phi,\zeta_1)\|_2$) and the next-state latent loss  ($\|\textsc{ENC}_{\phi_1}(I_{t}) - \hat{z}_{t}(\phi,\zeta_1)\|_2$). The model is then optimized with a weighted joint loss (with a hyperparameter $\beta_\textrm{Joint}$) as in Equation \eqref{J_rgb_Loss_eq}. Proper tuning of $\beta_\textrm{Joint}$ is crucial for balancing task-specific performance with model generalization.

\begin{subequations}
\begin{equation}
\begin{aligned}
 L_{\textrm{RGB}}(\phi_{1,2},\zeta_1) &= E_{D}\big[\|I_{t} - \hat{I}_{t}(\phi_{1,2},\zeta_1)  \|_2 + \\ &\beta_z \|\textsc{ENC}_{\phi_1}(I_{t}) - \hat{z}_{t}(\phi_{1,2})  \|_2\big]
 \end{aligned}
 \label{J_rgbLoss_eq}
\end{equation}
\begin{equation}
 L_\textrm{Joint}(\theta,\phi_{1,2},\zeta_1) =  E_{D}\big[ L_{\textrm{RGB}} +  \beta_\textrm{Joint} L_{\pi}\big] 
 \label{J_rgb_Loss_eq}
\end{equation}
\end{subequations}

\textbf{Decoupled Training.}
Inspired by reinforcement learning settings \cite{dreamerv3}\cite{ha2018worldmodels}, on contrast to joint training, we propose a decoupled training approach. The world model is initially pretrained using an unsupervised RGB reconstruction loss as in Equation \eqref{J_rgbLoss_eq}. Upon training, the world model's weights are frozen and then employed for learning various tasks that share dynamics to the training dataset $D$. To train the policy, the dataset is first processed to pre-compute the latent $z_t$ and the hidden state $h_t$. Subsequently, the policy $\pi_\theta$ is optimized with L1-loss, where only policy weights $\theta$ are optimized ($L_{\pi}(\theta) =  E_{D}\big[\|a_t - \hat{a}_t(\theta) \|\big]$). Such pretraining regime can leverage more diverse datasets enabling improved generalization. Compared to joint training, it might also mitigate the need for additional tuning of the RGB/policy loss weightings.   

\textbf{Supervised Dynamics Loss.}
Instead of solely relying on the unsupervised RGB loss to train the world model, we propose integrating a supervised dynamics prediction loss to obtain a dynamics-informed world model. The position, velocity, and acceleration-based losses are utilized either independently or in combination with other losses. Like the RGB reconstruction loss, the dynamics L2-loss is calculated as shown in Equation \eqref{dyn_WMLoss_eq}. The incorporation of such losses will encourage the world model to capture features relevant to dynamics within its latent embedding and hidden states. Consequently, better policies can be trained for different dynamic tasks. 

\begin{equation}
 L_{X\in[P,V,A]}(\phi,\zeta) = E_{D}\big[\|X_{t} - \hat{X}_{t}(\phi,\zeta)\|_2\big]
\label{dyn_WMLoss_eq}
\end{equation}

To investigate the significance of supervised dynamics mapping, it was introduced to both the joint and the decoupled training approaches. In the context of decoupled training, failure of the world model to capture task-relevant features would lead to poor performance. Thus, dynamics-informed models would be of greater use to the decoupled training scenarios for dynamic tasks. 

\section{Task Description}\label{task_desc}

\subsection{General Overview}
In this study, we initially focused on 2D tasks before progressing to 3D simulations or real-world tasks. The tasks were simulated using the Python PyMunk \cite{Pymunk} 2D physics simulation library. Given the simulation state, the visuals were rendered in PyGame \cite{Pygame}; an Open-GL based renderer.  

The visual inputs $I_t$ are 64$\times$64 RGB images. The action $a_t$ is three-dimensional, comprising relative position updates ($a_t=\Delta X_t,\Delta Y_t,\Delta \theta_t]$). The absolute position is pre-processed into an acceleration command ($a_{\textrm{acc}_t}=\frac{a_{\textrm{pos}_t}-2*a_{\textrm{pos}_{t-1}}+a_{\textrm{pos}_{t-2}}}{\Delta t^2}$) before sending to the PyMunk engine. The dynamic states ($[P_t,V_t,A_t]$) are a concatenation of the cart's and the block's $X,Y,\theta$ state (i.e.\ $P_t=[P_{c_x},P_{c_y},P_{c_\theta},P_{b_x},P_{b_y},P_{b_\theta}]$). The position and velocity states were directly provided by PyMunk, while the acceleration was differentially estimated ($A_t = \frac{V_t - V_{t-1}}{\Delta t}$). 

\begin{figure}[ht]
\centering
\includegraphics[width=\columnwidth]{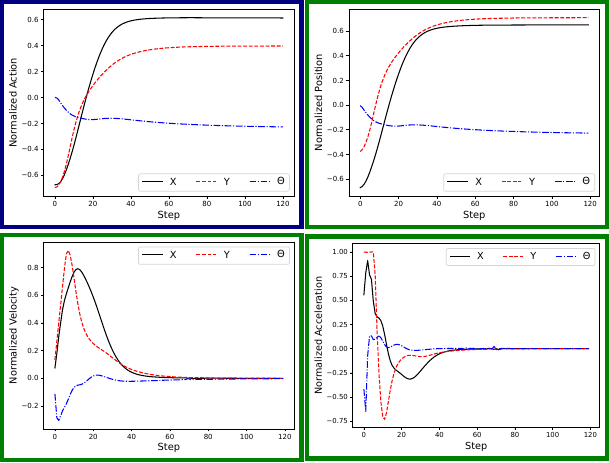}
\caption{A sample dataset trajectory for the ``Balance-Reaching" task) depicting the normalized position action (in blue) and different dynamics states (in green).}
\label{fig:DS_dyn}
\end{figure}

Figure \ref{fig:DS_dyn} illustrates the normalized (to [-1,1] range) position action (applied to the cart) and the dynamics states (of the block) of a sample trajectory. Position control was favored given its smoother trajectories, which are easier to imitate, as opposed to the rapidly changing acceleration state.

\subsection{Tasks Specification}
\begin{figure}[ht]
\centering
\begin{subfigure}{\columnwidth}
\includegraphics[width=\textwidth]{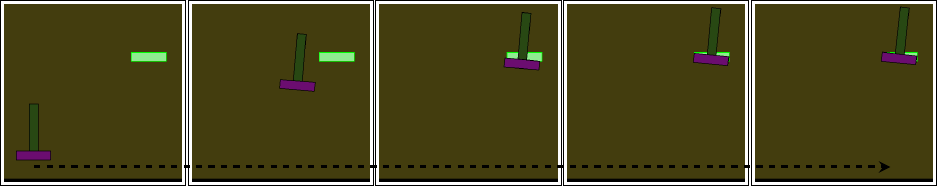}
\caption{\textbf{Task1:} Balance-Reaching (Core Task)}
\label{fig:BR_task}
\end{subfigure}
    \hfill
\begin{subfigure}{\columnwidth}
\includegraphics[width=\textwidth]{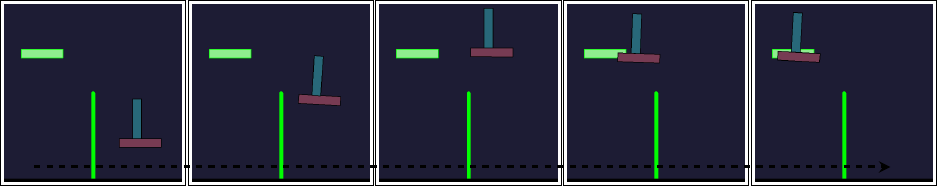}
\caption{\textbf{Task2:} Balance-Reaching[v2] (in-Domain Dynamics)}
\label{fig:inDomain}
\end{subfigure}
    \hfill
\begin{subfigure}{\columnwidth}
\includegraphics[width=\textwidth]{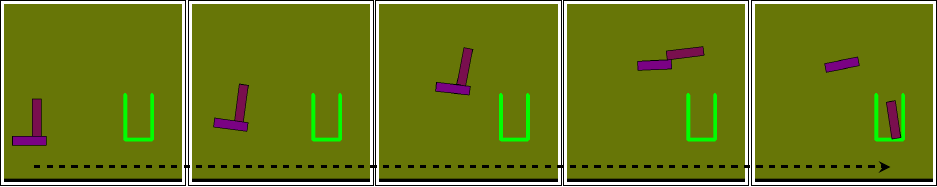}
\caption{\textbf{Task3:} Bin-Dropping (out-of-Domain Dynamics)}
\label{fig:outDomain}
\end{subfigure}
  \caption{Sample trajectories depicting the objective of different target tasks.}
\label{fig:task_desc} 
\end{figure}

A set of 2D tasks was designed within the same environment, utilizing identical simulation parameters. We had three main tasks as shown in Fig. \ref{fig:task_desc}. All tasks involve non-prehensile manipulation of a single block (initially vertical) placed on a position-controlled cart (initially horizontal), aiming towards a goal state (green).

\textbf{Task1: Balance-Reaching} (Fig. \ref{fig:BR_task}). The objective is for the cart to reach the target goal without the block falling off. In the absence of precise control, the block would lose balance and tip over. Moreover, careful acceleration control is necessary to stop at the goal within minimal time.

\textbf{Task2: Balance-Reaching[v2]} (Fig. \ref{fig:inDomain}). Mostly similar to task1, with the need to avoid an obstacle that is added at the center (green segment).

\textbf{Task3: Bin-Dropping} (Fig. \ref{fig:outDomain}). Contrary to block balancing objective, the goal here is to drop the block into the green bin by tipping it over, thereby altering the dynamics distribution.

\begin{figure}[ht]
\centering
\includegraphics[width=\columnwidth]{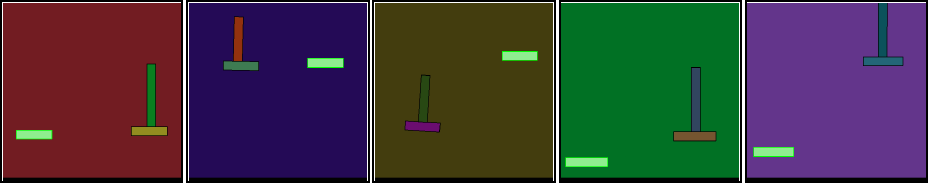}
\caption{Example dataset randomization  including start/target positions, the cart/block dimensions, and objects colors.}
\label{fig:DS_random}
\end{figure}

To increase task complexity and encourage improved generalization, the tasks were randomized in terms of the cart width, block height, cart/target XY position, and block shift around the cart center. The cart, block, and background colors are randomized acting as visual distractors. Figure \ref{fig:DS_random} illustrates randomization samples for task1.

\subsection{Expert Dataset Generation}
In the context of imitation learning, an expert dataset is necessary. For the dynamic tasks described, obtaining expert demonstrations from humans (i.e.\ via keyboard (discrete control) or 3D mouse (continuous control)) proved unfeasible. Objects were frequently dropped, and successful trajectories were sub-optimal (too slow) and unnatural (moving in one DOF at a time). Additionally, we could not identify simple scripted policies that can successfully complete those tasks. Consequently, we settled on training an expert DRL policy to generate the target dataset. 

\begin{figure}[ht]
\centering
\includegraphics[width=\columnwidth]{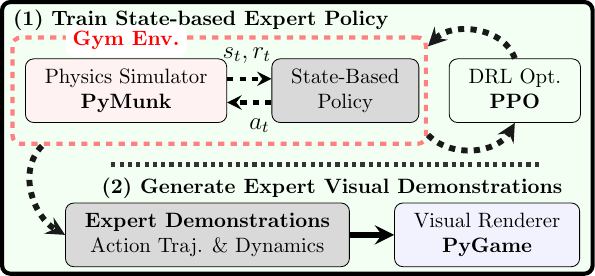}
\caption{Expert dataset generation utilizing a PPO-optimized state-based policy.}
\label{fig:DS_gen}
\end{figure}

First, we trained a set of expert policies based on a state-based reinforcement learning optimization algorithm as depicted in Fig. \ref{fig:DS_gen}. Compared to visual RL, State-based environments are typically easier to optimize. A PyMunk environment of the target task is first wrapped into a Gym environment \cite{gym}, specifying the state-based observation space (objects positions, velocities, and dimensions, as well as the target position), action space (cart acceleration), and task-specific reward (continuous for tasks 1 and 2, based on distance to goal, or sparse for task 3, based on whether the object is dropped or in the bin). The policy is then optimized via the PPO algorithm \cite{PPO}, utilizing the Stable-Baselines3  library \cite{SB3}. We produced a dataset of successful trajectories by employing various policy checkpoints, introducing multi-modal behavior. These trajectories were subsequently rendered in PyGame to obtain the visual RGB states. 

\section{Results \& Discussions}\label{results}
For all tasks, we generated 300 training trajectories and 50 evaluation trajectories. The tasks were simulated with a 1ms time-step ($\Delta t$) and controlled at a 20 Hz rate. The visual image (64$\times$64) represented a 1m$\times$1m space (1pix=15.625mm). Such discretization would limit the accuracy of dynamics prediction and goal reaching. The primary task, "Balance-Reaching," involved training different world model architectures and policies. For the other two tasks, we trained only the policy on top of frozen world models to assess model generalization. 

\subsection{Qualitative Evaluation (Task1)}
\begin{figure}[ht]
\centerline{\includegraphics[width=0.95\columnwidth]{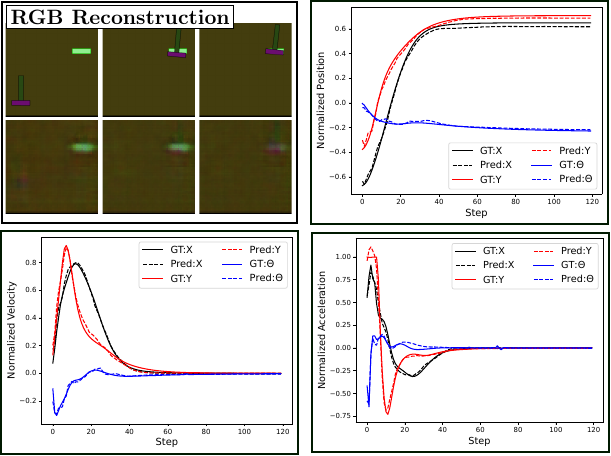}}
\caption{Qualitative evaluation of the image reconstruction and the dynamics prediction. \textbf{(Task1)}}
\label{fig:Recon}
\end{figure}

The world model was qualitatively evaluated by visualizing the RGB image reconstruction and the dynamics prediction as depicted in Fig. \ref{fig:Recon}. The RGB reconstruction fairly captured the objects' colors and their approximate positions. However, it struggled to accurately represent the objects' dimensions, relative position, and angular positions. As a result, relying solely on RGB might limit performance on dynamic tasks. Regarding dynamics, the model predicted slowly varying position states with higher accuracy compared to velocity or rapidly-changing acceleration states. The inaccuracy of goal reconstruction suggested the need for direct goal position incorporation into the policy. 

\begin{figure}[ht]
\centering
\begin{subfigure}{0.49\columnwidth}
\includegraphics[width=\textwidth]{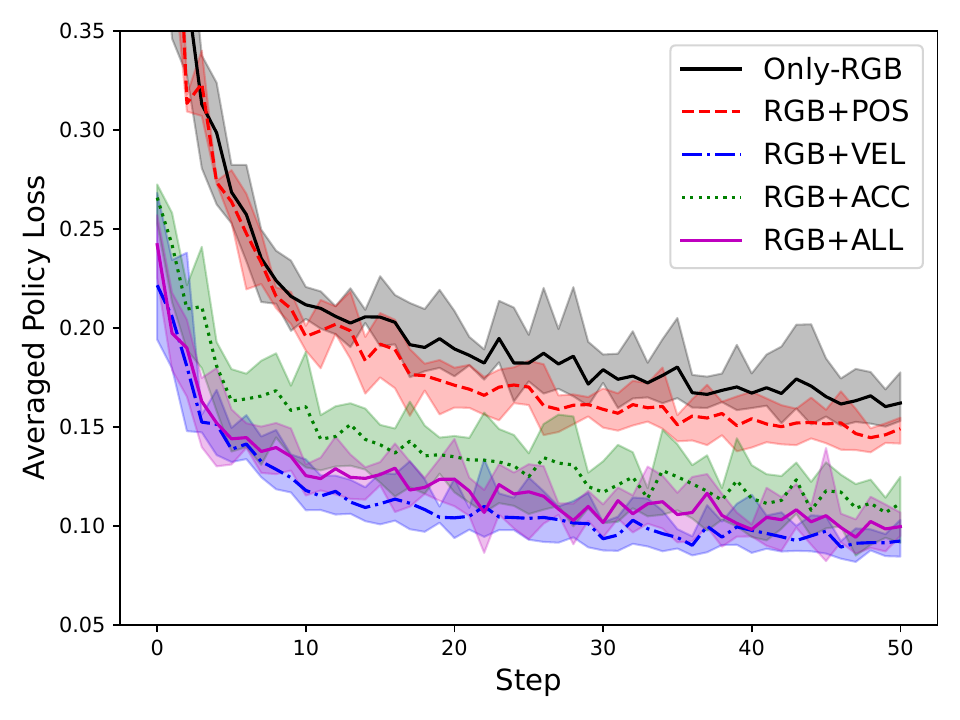}
\caption{Dynamics Supervision}
\label{fig:learning_dyn}
\end{subfigure}
\hfill
\begin{subfigure}{0.49\columnwidth}
\includegraphics[width=\textwidth]{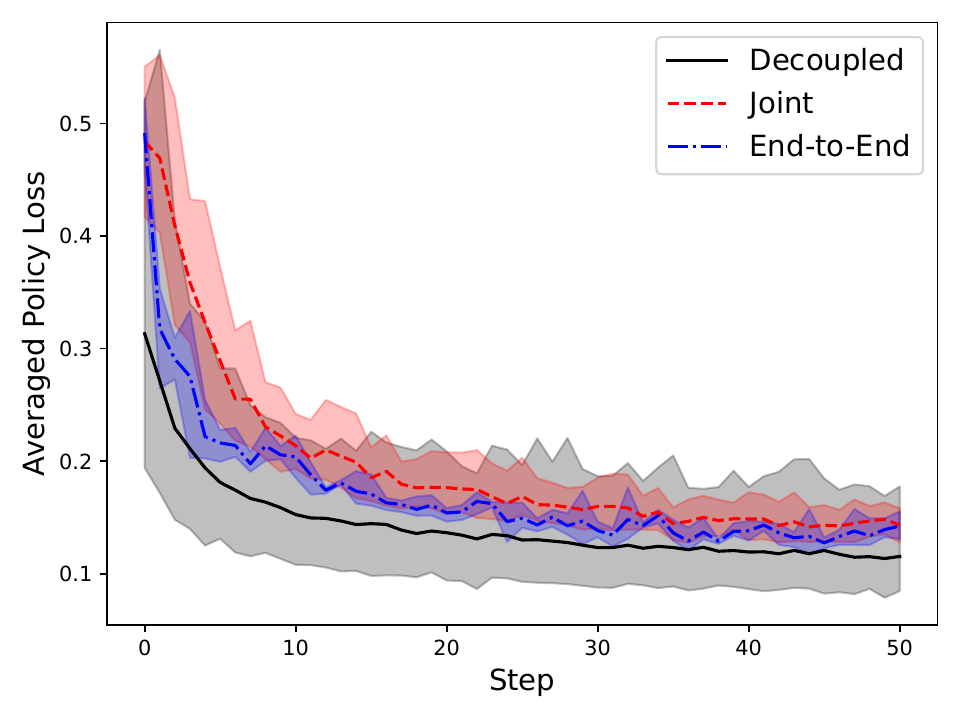}
\caption{Training Approach}
\label{fig:learning_training}
\end{subfigure}
\begin{subfigure}{0.49\columnwidth}
\includegraphics[width=\textwidth]{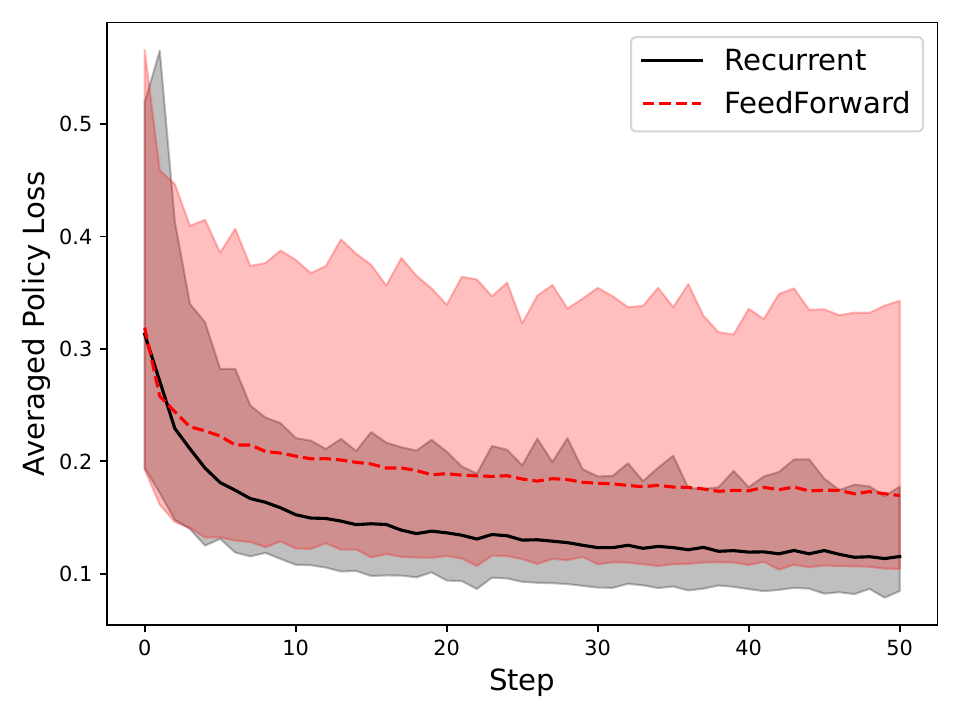}
\caption{Policy Type}
\label{fig:learning_rec}
\end{subfigure}
\hfill
\begin{subfigure}{0.49\columnwidth}
\includegraphics[width=\textwidth]{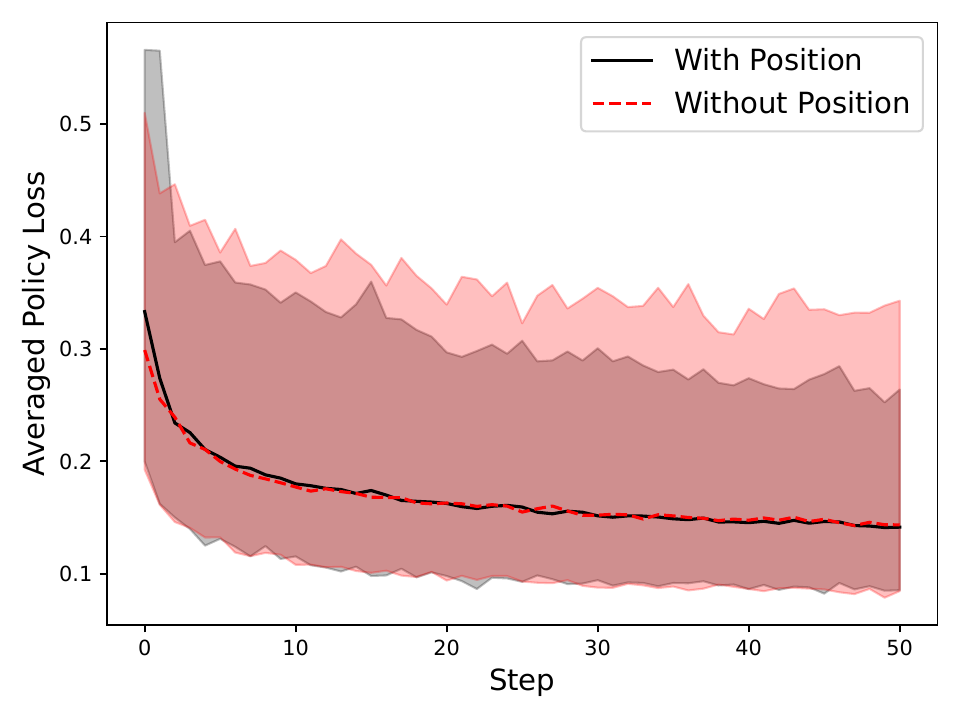}
\caption{Position Prediction}
\label{fig:learning_pos}
\end{subfigure}
\begin{subfigure}{0.49\columnwidth}
\includegraphics[width=\textwidth]{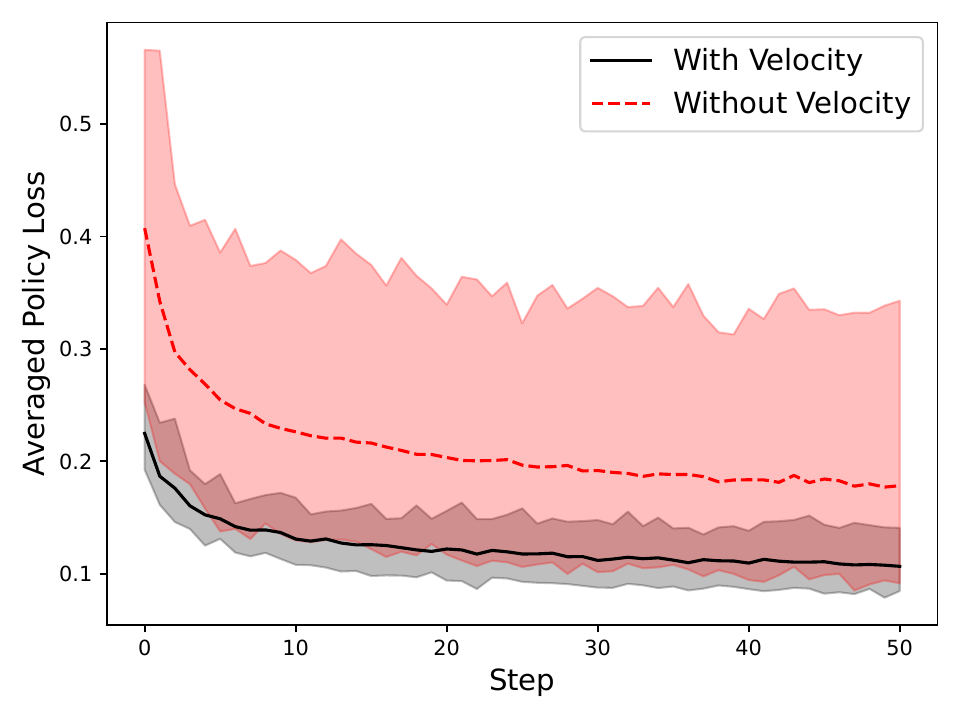}
\caption{Velocity Prediction}
\label{fig:learning_vel}
\end{subfigure}
\begin{subfigure}{0.49\columnwidth}
\includegraphics[width=\textwidth]{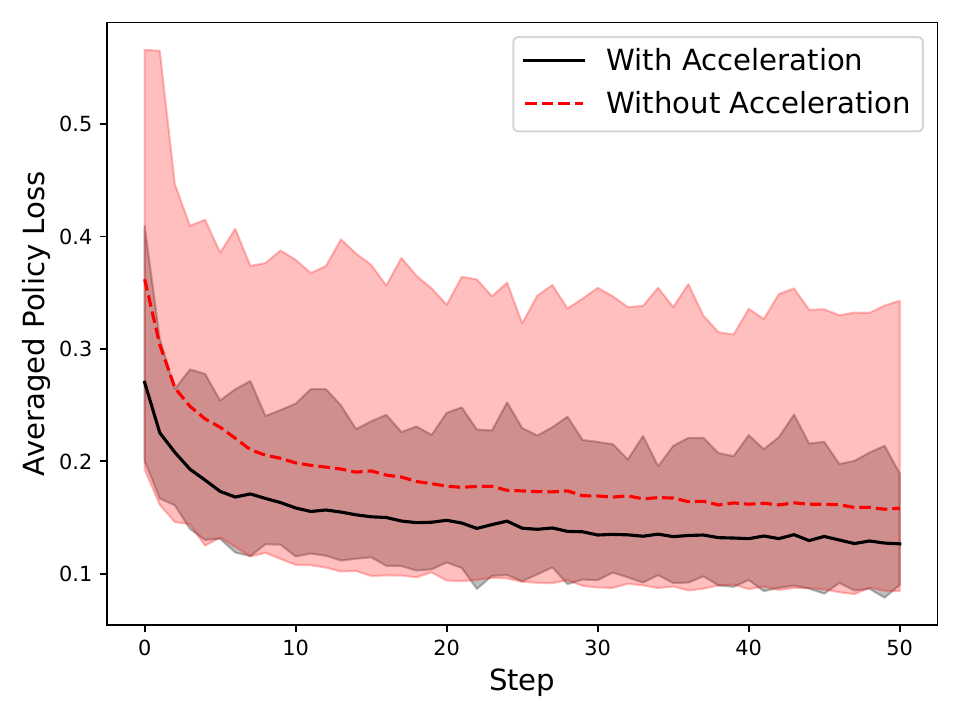}
\caption{Acceleration Prediction}
\label{fig:learning_acc}
\end{subfigure}
\caption{\textbf{(Task1)} Policy loss comparison of different model architectures illustrating the utility of dynamics supervision, decoupled training, and recurrent policy.}
\label{fig:learning} 
\end{figure}

Prior to quantitative evaluation, the policy loss can serve as a strong indicator of the model performance by depicting the general tendencies over multiple runs and architectures. In Fig. \ref{fig:learning_dyn}, under decoupled training, dynamics-informed models had lower losses than ``Only-RGB" models as shown . The velocity state was the most effective when compared to position—offering limited dynamic information—or acceleration, which varies too rapidly. Combining them only increased the loss relative to the velocity-based model. In Fig. \ref{fig:learning_training}), decoupled training realized lower losses compared to joint or end-to-end training thanks to limited over-fitting and the need for loss weight fine-tuning ($\beta_\textrm{Joint}$). In Fig. \ref{fig:learning_rec}, recurrent policies had lower losses compared to feedforward ones as they can handle non-Markovian transitions. Incorporation of different dynamical states had varying effects where position (Fig.\ref{fig:learning_pos}) was the least effective, and velocity (Fig. \ref{fig:learning_vel}) had the most notable improvement.

\subsection{Quantitative Evaluation (Task1)}
We evaluated the performance for task1 (``Balance-Reaching"). The evaluation dataset comprised 50 randomized trajectories (lasting 120 steps each). Each model was evaluated with three random seeds.

\begin{figure}[ht]
\centering
\begin{subfigure}{0.49\columnwidth}
\includegraphics[width=\textwidth]{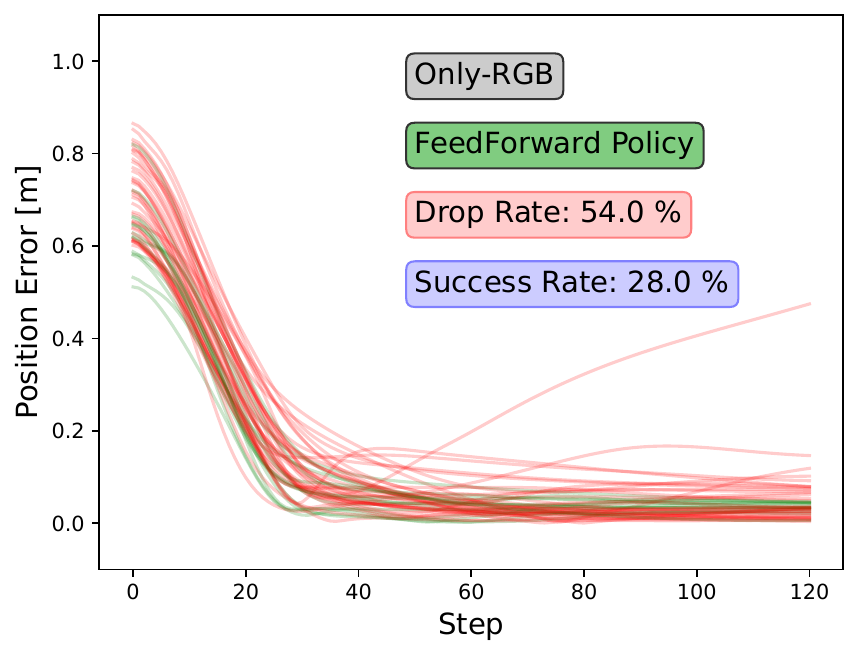}
\caption{Only-RGB}
\label{fig:eval_rgb}
\end{subfigure}
    \hfill
\begin{subfigure}{0.49\columnwidth}
\includegraphics[width=\textwidth]{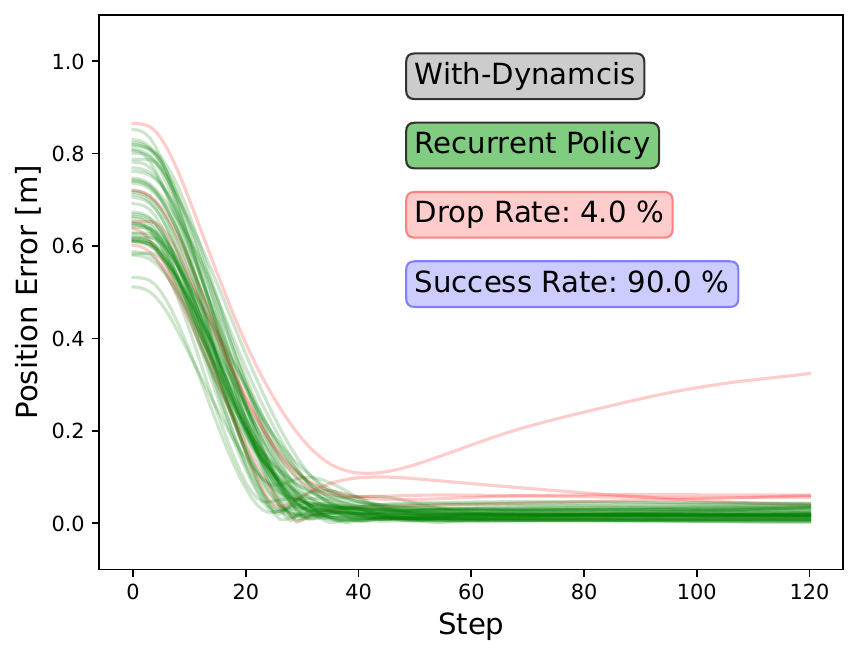}
\caption{With Dynamics}
\label{fig:eval_dyn}
\end{subfigure}
  \caption{Visualization of the position error trajectory with/without dynamics mapping. Red color represents failure due to dropped block. \textbf{(Task1)}}
\label{fig:evalCriteria} 
\end{figure}

\textbf{Evaluation Criteria.} Figure \ref{fig:evalCriteria} illustrates the evaluation criteria comparing two models under decoupled training. The first model is an Only-RGB model with feedforward policy, and the second utilizes dynamics-supervision and a recurrent policy. The first criterion is balancing the block throughout the trajectory. The ``Drop Rate" refers to the percentage failures due to block drops. The second criterion is mean position error over the last 10 steps. A trajectory qualifies as successful if the block remains in place and the final error is under 50 mm (approximately 3 pixels). Introducing dynamics (Fig. \ref{fig:eval_dyn}) improves performance over ``Only-RGB" (Fig. \ref{fig:eval_rgb}) by reducing both the drop rate and position error. 

\textbf{Decoupled Training.} We compared a world model trained with an ``Only-RGB" reconstruction loss against dynamics-informed models that include position (P), velocity (V), acceleration (A), or their combinations (with equal loss weightings). Given our task1 dataset, we trained a single world model (per architecture) as well as three randomly seeded policies. Table \ref{tab:T1_dec} compares the drop rate (DR), the final position error (PE), and the success rate (SR) across models. 

With a feedforward policy, the ``Only-RGB" model exhibited a high drop rate resulting in an overall lower success rate. Its position error was also higher on average. The model policy, trained on poor latent features without dynamics, failed to learn appropriate multi-modal behavior and frequently went out of distribution. Introducing a recurrent policy can significantly improve the ``Only-RGB" performance as it learns necessary transition dynamics during policy learning.

With feedforward policies, introducing position (P) prediction had a slightly improved performance. Yet, due to its static, non-temporal nature, relatively high drop rates were noted. The best performance was attained by incorporating velocity, resulting in the lowest DR and PE as well as the highest SR. Compared to acceleration, velocity trajectories were smoother and easier to predict (Fig. \ref{fig:Recon}). Moreover, ground-truth acceleration was differentially estimated, whereas velocity was directly reported. Consequently, acceleration-based models had a relative increase in DR and PE. The combination of position/velocity improved over their standalone performance by reducing the drop rates, and position error. While the velocity addressed the out-of-distribution states causing object drop, the position improved the goal tracking performance. Other combinations (e.g.\ P+V+A) had comparable performance. Similar tendencies can be noted with recurrent policies with an overall higher performance. Dynamics-supervision is less significant since necessary dynamics transition are learnt by the recurrent policy to optimize policy loss.

\begin{table*}[ht]
\centering
\caption{\textbf{Base Task -- Task1.} Comparison of different models performance w \& w/o dynamics.}
\begin{threeparttable}
\centering
{\begin{tabular}{l|ccc|ccc|ccc}
 {} &  \multicolumn{3}{c}{\textbf{Decoupled [FeedForward]}} &  \multicolumn{3}{c}{\textbf{Decoupled [Recurrent]}}  &  \multicolumn{3}{c}{\textbf{Joint [Recurrent]}} \\ \hline \hline 

\textbf{Architecture} &  \textbf{DR}\textsuperscript{a}[\%] $\downarrow$ &\textbf{PE}[mm] $\downarrow$ & \textbf{SR}[\%] $\uparrow$ & \textbf{DR}[\%] $\downarrow$ &\textbf{PE}[mm] $\downarrow$ & \textbf{SR}[\%] $\uparrow$  & \textbf{DR}[\%] $\downarrow$ &\textbf{PE}[mm] $\downarrow$ & \textbf{SR}[\%] $\uparrow$ \vspace{2pt} \\ \hline \hline 
    End-to-End &   - & - & - & - & - & - & 19 (9.0) & 71 (21.9) & 57 (10.6)\\ 
    Only-RGB &   55 (2.5) & 69 (15.6) & 21 (5.7) & 14 (1.6) & 31 (1.3) & 73 (3.8) & 17 (5.0) & 92 (31.4) & 64 (5.9)\\ \hline 
      RGB+[P] &  31 (8.4) & 33 (2.8) & 62 (9.1) & 4 (3.3) & \textbf{21 (2.7)} & \textbf{92 (4.3)} & \textbf{5 (0.9)} & 58 (17.1) & 67 (8.1) \\
      RGB+[V] &   11 (2.5) & 41 (5.8) & 79 (1.9) & 7 (1.9) & 29 (5.0) & 85 (3.8) & 22 (6.5) & 84 (31.6) & 51 (10.9) \\
      RGB+[A] &  15 (3.4) & 80 (21.2) & 53 (5.7) & 5 (0.9) & 36 (3.5) & 79 (4.1) & 15 (3.8) & 43 (4.1) & 65 (3.8) \\
    RGB+[P+V] &   \textbf{7 (5.0)} & \textbf{25 (2.2)} & \textbf{85 (5.2)} & 5 (2.5) & 23 (3.7) & 89 (0.9) & 9 (1.9) & 54 (11.0) & 59 (9.6) \\
    RGB+[P+A] &  34 (4.3) & 56 (4.7) & 42 (8.6) & \textbf{2 (1.6)} & 32 (6.1) & 84 (1.6) & 6 (4.3) & \textbf{34 (2.1)} & \textbf{79 (6.2)}\\
    RGB+[V+A] &  9 (2.5) & 31 (7.0) & 80 (1.6) & 8 (4.3) & 28 (7.0) & 82 (7.1)  & 19 (0.9) & 48 (6.1) & 57 (5.7)  \\
  RGB+[P+V+A] &   9 (5.0) & 25 (1.6) & 84 (7.5)  & 5 (0.9) & 25 (2.2) & 87 (1.9) & 7 (2.5) & 43 (7.2) & 70 (8.6)\\ \hline
\end{tabular}}
\begin{tablenotes}
  \small
  \item \textsuperscript{a}DR = ``Drop Rate", PE = ``Position Error", and SR = ``Success Rate".
\end{tablenotes}
\end{threeparttable}
\label{tab:T1_dec}
\end{table*}

\textbf{Joint Training.} We evaluated different models over three random seeds. The policy loss is optimized jointly with the world model losses ($\beta_\textrm{Joint}=1$). The ``End-to-End" model employed only the policy loss to optimize both modules. As listed in Table \ref{tab:T1_dec}, ``End-to-End" and ``Only-RGB" models (no dynamics) had improved performance over their decoupled training counterparts. The ``End-to-End" model had marginally lower performance over the ``Only-RGB" model due to potential over-fitting. Compared to decoupled training, the joint approach performance dropped slightly. Carefully tuning the loss weight hyperparameter $\beta_\textrm{Joint}$ can improve the performance without compromising generalization. Nevertheless, like decoupled training, improved performance is generally noted upon introducing dynamics losses. 

\textbf{Note on Real-World Applications:} After verifying the benefits of dynamics for improved performance in simulated settings, extension to real-world applications is a natural next objective. Considering the challenges of acquiring ground-truth dynamics states in real-world settings, we posit that models trained in simulation—addressing the sim-to-real gap effectively—can be zero-shot transferred to real-world settings\cite{hanai_force_2023}. During real-world inference, task-relevant dynamics will be implicitly encoded into the model's latent solely from the visual input, eliminating the need for explicit ground-truth dynamics.

\subsection{World Model Generalization}
The dynamics-informed models demonstrated an improved performance when evaluated on same task as the training dataset. Furthermore, we would like to verify whether these models exhibit good generalization behavior. To this end, we reuse a frozen world model, trained on task1, to train recurrent policies (3 seeds) for two additional tasks. Task2, while experiencing similar dynamics (in-domain), had a slightly different objective. Conversely, task3 had quite different dynamics (out-of-domain), due to increased angular positions and velocities.

\textbf{In-Domain Generalization -- Task2} The results presented in Table \ref{tab:inDomain} indicate a good transfer of the learnt model to the new task. Unlike expectations, task2 dataset was better imitated resulting in higher success rates. Similar to task1, dynamics-supervision and decoupled training advantage were observed.  

\begin{table}[ht]
\caption{\textbf{In-Domain Generalization -- Task2.} Comparison of different models performance w \& w/o dynamics.}
\centering
{\begin{tabular}{l|cc|cc}
 {} &  \multicolumn{2}{c}{\textbf{Decoupled}} &   \multicolumn{2}{c}{\textbf{Joint}} \\
\textbf{Arch.} &  \textbf{DR}[\%] $\downarrow$ & \textbf{SR}[\%] $\uparrow$ & \textbf{DR}[\%] $\downarrow$ & \textbf{SR}[\%] $\uparrow$ \vspace{2pt} \\ \hline \hline 
    End-to-End&    -     &  -      &  4 (1.6) & 72 (4.3) \\
    Only-RGB &   1 (0.9) & 80 (5.9)           & 4 (1.6) & 73 (4.7)\\ \cmidrule{1-5} 
    +[P]     & 0 (0.0)   & \textbf{91 (3.8)}  & 1 (0.9) & \textbf{86 (4.9)} \\
    +[V]     &  2 (1.6)  & 80 (4.3)           & 3 (2.5) & 79 (6.6)\\
    +[A]     &  2 (1.6)  & 82 (4.9)           & 3 (2.5) & 77 (6.6)\\ 
    +[P+V]   &  1 (0.9)  & \textbf{90 (3.3)}  & 4 (3.3) & 75 (6.2) \\
    +[P+A]   &   0 (0.0) & 87 (2.5)           & \textbf{0 (0.0)} & 76 (7.1) \\
    +[V+A]   &   1 (0.9) & 83 (8.1)           & 3 (0.9) & 83 (4.1)\\
    +[P+V+A] &   2 (1.6) & 84 (3.3)           & 1 (1.9) & \textbf{87 (6.6)} \\ \hline
\end{tabular}}
\label{tab:inDomain}
\end{table}

\textbf{Out-of-Domain Generalization -- Task3} For this task, the success rate (block dropped in bin) is the only relevant metric. As shown in Table \ref{tab:outDomain}, for out-of-domain tasks, dynamics-supervision with decoupled training had limited improvement in performance compared to ``Only-RGB" models. On the other hand, under joint training, dynamics-supervision had marginally lower performance,

\begin{table}[t]
\caption{\textbf{Out-of-Domain Generalization -- Task3.} Comparison of different models performance w \& w/o dynamics.}
\centering
{\begin{tabular}{l|c|c}
\textbf{Arch.} & \textbf{Decoupled} (SR[\%]) & \textbf{Joint} (SR[\%]) \vspace{2pt} \\ \hline \hline 
    End-to-End &  - & 47 (4.1) \\
    Only-RGB &   44 (0.0) & 46 (1.6) \\ \cmidrule{1-3} 
    +[P]     & 46 (4.3)   & \textbf{51 (4.1)} \\
    +[V]     &  52 (1.6)  & 40 (2.8) \\
    +[A]     &  \textbf{55 (6.2)}  & 41 (12.3) \\ 
    +[P+V]   &  53 (3.4)  & 42 (8.2) \\
    +[P+A]   &   46 (2.8) & 39 (4.1) \\
    +[V+A]   &   41 (3.4) & 46 (9.1) \\
    +[P+V+A] &   \textbf{54 (2.8)} & 45 (5.2) \\ \hline
\end{tabular}}
\label{tab:outDomain}
\end{table}

\begin{figure}[ht]
\centerline{\includegraphics[width=0.9\columnwidth]{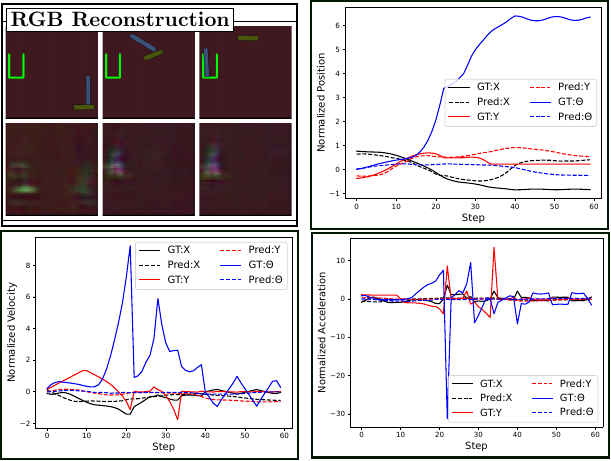}}
\caption{Qualitative evaluation of the out-of-domain image reconstruction and
the dynamics prediction. (Task3)}
\label{fig:Recon_T2}
\end{figure}

The suboptimal performance of dynamics-based models can be visually explained in Fig. \ref{fig:Recon_T2}. These models, including the "Only-RGB" model, had poor RGB reconstruction, failing to accurately capture the cart/block pose and the goal's shape and position. Such behavior is influenced by both the world model (encoded state) and the decoder. Moreover, poor dynamics predictions in cases of extreme angular states suggest out-of-distribution (OOD) behavior, leading to OOD latent features and hidden states that pose challenges for both learning and inference.

\section{CONCLUSIONS}
In this study, we propose the incorporation of supervised dynamics prediction losses to learn dynamics-informed world models. In our simplified non-prehensile manipulation tasks, the proposed models exhibited lower policy losses and enhanced performance for both decoupled and joint training regimes. Generalizing these world models to additional tasks suggests the importance of avoiding out-of-distribution states for transferable performance.

\textbf{Limitations \& Future Work} The accuracy of vision-based dynamics inference is dependent on image size. For different tasks, a compromise between accuracy and computational efficiency necessitates consideration of varying image sizes. Moreover, the precision of predictions is bounded by the ground-truth training dataset's accuracy. Better simulation engines and reduced simulation and control time steps may be required for certain applications. Hyperparameter tuning represents another investigation direction to assess the robustness of our proposal. In addition to the baselines presented, further comparisons with SOTA baselines—namely, Diffusion Policy \cite{chi2023diffusionpolicy} and Dreamer \cite{dreamerv3}—are planned.
In assessing our proposal, we opted for simple neural network architectures, including vanilla CNNs and LSTMs. A CNN model proved adequate for encoding/decoding the current simple visuals, while LSTMs were chosen for their effective online performance in tasks with extended horizons. Future research will incorporate more recent models for both visual (e.g., ResNet, ViT) and sequential (e.g., Transformers, Diffusion Models) processing. Investigations will also extend to 3D simulated and real-world environments to explore sim2real generalization. Moreover, examining tasks involving multiple manipulated objects presents a promising direction for generalizing to object-centric models (e.g., GNN\cite{Ugadiarov2023GraphicalOA}). Lastly, we seek models that learns directly from image/action pairs without reliance on specialized dynamics datasets. Optical flow models (e.g., Flowformer \cite{huang2022flowformer}) can potentially estimate target scene dynamics.


\section*{ACKNOWLEDGMENT}
The manuscript was reviewed for linguistic accuracy with the assistance of GPT-4. This work was supported by JST [Moonshot R\&D][Grant Number JPMJMS2031].


\bibliographystyle{IEEEtran}
\bibliography{root}

\begin{thebibliography}{10}
\providecommand{\url}[1]{#1}
\csname url@samestyle\endcsname
\providecommand{\newblock}{\relax}
\providecommand{\bibinfo}[2]{#2}
\providecommand{\BIBentrySTDinterwordspacing}{\spaceskip=0pt\relax}
\providecommand{\BIBentryALTinterwordstretchfactor}{4}
\providecommand{\BIBentryALTinterwordspacing}{\spaceskip=\fontdimen2\font plus
\BIBentryALTinterwordstretchfactor\fontdimen3\font minus \fontdimen4\font\relax}
\providecommand{\BIBforeignlanguage}[2]{{%
\expandafter\ifx\csname l@#1\endcsname\relax
\typeout{** WARNING: IEEEtran.bst: No hyphenation pattern has been}%
\typeout{** loaded for the language `#1'. Using the pattern for}%
\typeout{** the default language instead.}%
\else
\language=\csname l@#1\endcsname
\fi
#2}}
\providecommand{\BIBdecl}{\relax}
\BIBdecl

\bibitem{nonprehensileSurvey}
F.~Ruggiero, V.~Lippiello, and B.~Siciliano, ``Nonprehensile dynamic manipulation: A survey,'' \emph{IEEE Robotics and Automation Letters}, vol.~3, no.~3, pp. 1711--1718, 2018.

\bibitem{quasiStatic}
S.~Young, D.~Gandhi, S.~Tulsiani, A.~Gupta, P.~Abbeel, and L.~Pinto, ``Visual imitation made easy,'' in \emph{Conference on Robot Learning (CORL)}, ser. Proceedings of Machine Learning Research, J.~Kober, F.~Ramos, and C.~Tomlin, Eds., vol. 155.\hskip 1em plus 0.5em minus 0.4em\relax PMLR, 16--18 Nov 2021, pp. 1992--2005.

\bibitem{keepitupright}
A.~Heins and A.~P. Schoellig, ``Keep it upright: Model predictive control for nonprehensile object transportation with obstacle avoidance on a mobile manipulator,'' \emph{IEEE Robotics and Automation Letters}, vol.~8, no.~12, pp. 7986--7993, 2023.

\bibitem{nonprehensileICRA}
J.~Z. Woodruff and K.~M. Lynch, ``Planning and control for dynamic, nonprehensile, and hybrid manipulation tasks,'' in \emph{2017 IEEE International Conference on Robotics and Automation (ICRA)}, 2017, pp. 4066--4073.

\bibitem{chi2024universal}
C.~Chi, Z.~Xu, C.~Pan, E.~Cousineau, B.~Burchfiel, S.~Feng, R.~Tedrake, and S.~Song, ``Universal manipulation interface: In-the-wild robot teaching without in-the-wild robots,'' in \emph{arXiv}, 2024.

\bibitem{chi2023diffusionpolicy}
C.~Chi, S.~Feng, Y.~Du, Z.~Xu, E.~Cousineau, B.~Burchfiel, and S.~Song, ``Diffusion policy: Visuomotor policy learning via action diffusion,'' in \emph{Proceedings of Robotics: Science and Systems (RSS)}, 2023.

\bibitem{zhao2023learning}
T.~Z. Zhao, V.~Kumar, S.~Levine, and C.~Finn, ``Learning fine-grained bimanual manipulation with low-cost hardware,'' in \emph{2023 Robotics: Science and Systems (RSS)}, 2023.

\bibitem{RL}
R.~S. Sutton and A.~G. Barto, \emph{Reinforcement Learning: An Introduction}.\hskip 1em plus 0.5em minus 0.4em\relax Cambridge, MA, USA: A Bradford Book, 2018.

\bibitem{levine_end--end_2016}
S.~Levine, C.~Finn, T.~Darrell, and P.~Abbeel, ``End-to-end training of deep visuomotor policies,'' \emph{The Journal of Machine Learning Research}, vol.~17, no.~1, pp. 1334--1373, Jan. 2016.

\bibitem{suzuki2023deep}
K.~Suzuki, H.~Ito, T.~Yamada, K.~Kase, and T.~Ogata, ``Deep predictive learning : Motion learning concept inspired by cognitive robotics,'' 2023.

\bibitem{ha2018worldmodels}
D.~Ha and J.~Schmidhuber, ``Recurrent world models facilitate policy evolution,'' in \emph{Advances in Neural Information Processing Systems (NIPS)}, 2018.

\bibitem{embed2control}
M.~Watter, J.~Springenberg, J.~Boedecker, and M.~Riedmiller, ``Embed to {Control}: {A} {Locally} {Linear} {Latent} {Dynamics} {Model} for {Control} from {Raw} {Images},'' in \emph{Advances in {Neural} {Information} {Processing} {Systems} (NIPS)}, 2015.

\bibitem{dreamerv3}
D.~Hafner, J.~Pasukonis, J.~Ba, and T.~Lillicrap, ``Mastering diverse domains through world models,'' \emph{arXiv preprint arXiv:2301.04104}, 2023.

\bibitem{rt12022arxiv}
A.~{Brohan, et al.}, ``Rt-1: Robotics transformer for real-world control at scale,'' in \emph{Robotics: Science and Systems (RSS)}, 2023.

\bibitem{hu2023gaia1}
A.~Hu, L.~Russell, H.~Yeo, Z.~Murez, G.~Fedoseev, A.~Kendall, J.~Shotton, and G.~Corrado, ``Gaia-1: A generative world model for autonomous driving,'' 2023.

\bibitem{demoss2023ditto}
B.~DeMoss, P.~Duckworth, N.~Hawes, and I.~Posner, ``Ditto: Offline imitation learning with world models,'' 2023.

\bibitem{Pymunk}
\BIBentryALTinterwordspacing
V.~Blomqvist, ``Pymunk, a python 2d physics library,'' Nov. 2007--2024. [Online]. Available: \url{https://pymunk.org}
\BIBentrySTDinterwordspacing

\bibitem{Pygame}
\BIBentryALTinterwordspacing
P.~Shinners, ``Pygame, a python modules designed for writing video games.'' 2011--2024. [Online]. Available: \url{https://pygame.org}
\BIBentrySTDinterwordspacing

\bibitem{gym}
\BIBentryALTinterwordspacing
M.~Towers, J.~K. Terry, A.~Kwiatkowski, J.~U. Balis, G.~d. Cola, T.~Deleu, M.~Goulão, A.~Kallinteris, A.~KG, M.~Krimmel, R.~Perez-Vicente, A.~Pierré, S.~Schulhoff, J.~J. Tai, A.~T.~J. Shen, and O.~G. Younis, ``Gymnasium,'' Mar. 2023. [Online]. Available: \url{https://zenodo.org/record/8127025}
\BIBentrySTDinterwordspacing

\bibitem{PPO}
J.~Schulman, F.~Wolski, P.~Dhariwal, A.~Radford, and O.~Klimov, ``Proximal policy optimization algorithms,'' \emph{ArXiv}, vol. abs/1707.06347, 2017.

\bibitem{SB3}
\BIBentryALTinterwordspacing
A.~Raffin, A.~Hill, A.~Gleave, A.~Kanervisto, M.~Ernestus, and N.~Dormann, ``Stable-baselines3: Reliable reinforcement learning implementations,'' \emph{Journal of Machine Learning Research}, vol.~22, no. 268, pp. 1--8, 2021. [Online]. Available: \url{http://jmlr.org/papers/v22/20-1364.html}
\BIBentrySTDinterwordspacing

\bibitem{hanai_force_2023}
R.~Hanai, Y.~Domae, I.~G. Ramirez-Alpizar, B.~Leme, and T.~Ogata, ``Force {Map}: {Learning} to {Predict} {Contact} {Force} {Distribution} from {Vision},'' in \emph{2023 IEEE/RSJ International Conference on Intelligent Robots and Systems (IROS)}, 2023.

\bibitem{Ugadiarov2023GraphicalOA}
L.~Ugadiarov and A.~I. Panov, ``Graphical object-centric actor-critic,'' \emph{ArXiv}, vol. abs/2310.17178, 2023.

\bibitem{huang2022flowformer}
Z.~Huang, X.~Shi, C.~Zhang, Q.~Wang, K.~C. Cheung, H.~Qin, J.~Dai, and H.~Li, ``{FlowFormer}: A transformer architecture for optical flow,'' \emph{{ECCV}}, 2022.

\end{thebibliography}

\addtolength{\textheight}{-12cm}   
                                  
\end{document}